\documentclass[10pt,twocolumn,letterpaper]{article}

\usepackage{cvpr}
\usepackage{times}
\usepackage{epsfig}
\usepackage{graphicx}
\usepackage{amsmath}
\usepackage{amssymb}
\usepackage{bm}
\usepackage{soul}
\usepackage{url}
\usepackage{algorithmic}
\usepackage{multirow}
\usepackage{tabularx}
\usepackage{xcolor}
\usepackage{stfloats}
\usepackage{caption}
\usepackage{siunitx}
\usepackage{flushend}
\usepackage{caption}

\usepackage[pagebackref=true,breaklinks=true,letterpaper=true,colorlinks,bookmarks=false]{hyperref}

\cvprfinalcopy 

\def\cvprPaperID{216} 

\ifcvprfinal\fi
\begin{document}

\title{3D Face Reconstruction from A Single Image Assisted by 2D Face Images in the Wild}

\author{\normalsize{Xiaoguang~Tu$^{1}$\footnote{Contact Author}, Jian~Zhao$^{2,3}$, Zihang~Jiang$^{4}$, Yao~Luo$^{1}$, Mei~Xie$^{1}$, Yang~Zhao$^{3}$, Linxiao~He$^{5}$, Zheng~Ma$^{1}$ and Jiashi~Feng$^{2}$} \\
	\small{$^{1}$University of Electronic Science and Technology of China, $^{2}$National University of Singapore}, \\
		\small{$^{3}$National University of Defense Technology}, \small{$^{4}$ University of Science and Technology of China}, \small{$^{5}$CASIA}. \\
	\small{xguangtu@outlook.com, zhaojian90@u.nus.edu, jzh0103@mail.ustc.edu.cn, 13509184045@163.com, mxie@uestc.edu.cn}, \\ \small{ zhaoyang10@nudt.edu.cn, lingxiao.he@nlpr.ia.ac.cn, zma\underline{ }uestc@outlook.com} , {\small elefjia@nus.edu.sg}}

\twocolumn[{%
\renewcommand\twocolumn[1][]{#1}%
\maketitle
\begin{center}
    \centering
    \includegraphics[width=17.1cm, height=7.4cm]{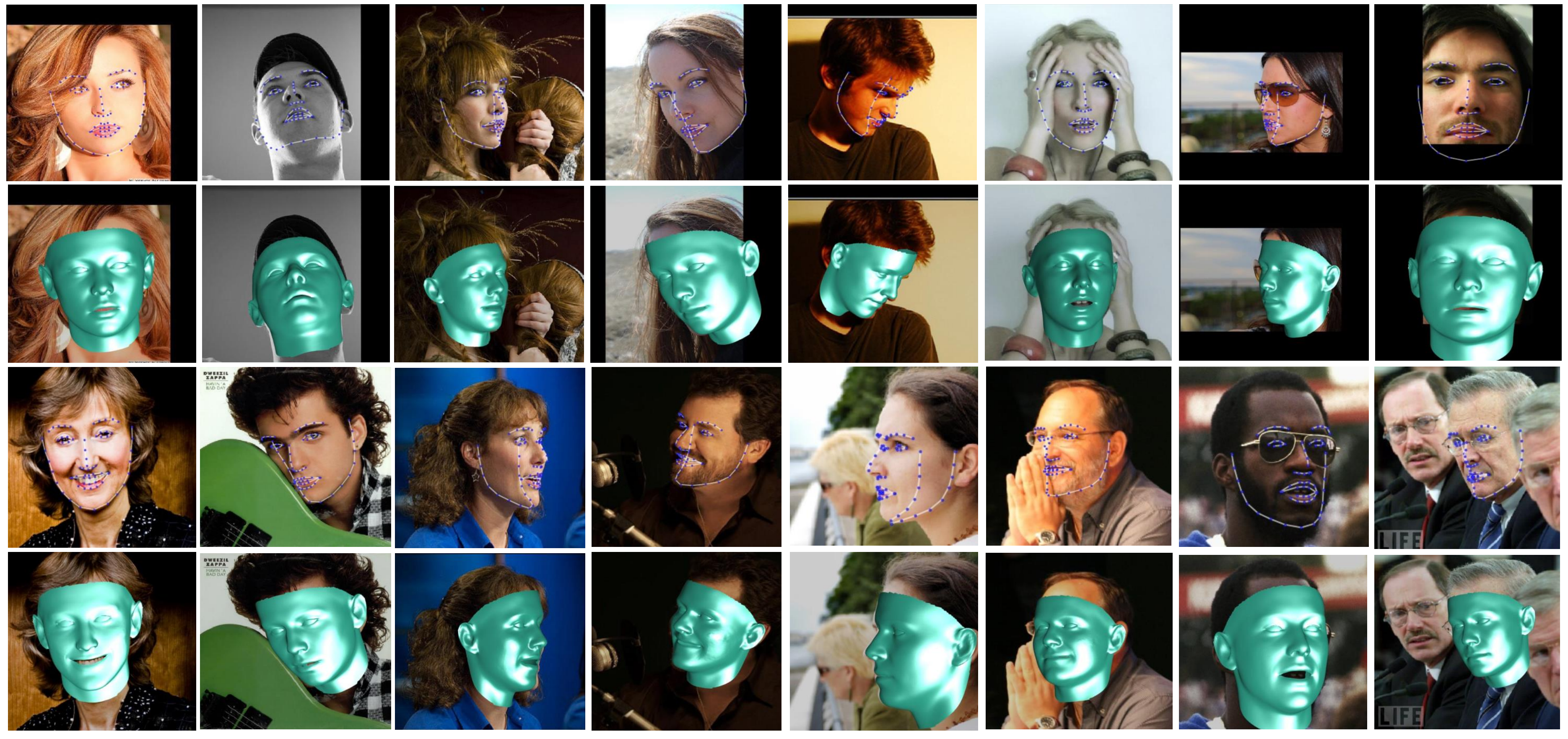}
    \vspace{-0.3cm}
    \captionsetup{font={small}}
    \captionof{figure}{{{Dense face alignment (odd rows) and 3D face reconstruction (even rows) results from our proposed  method. For alignment, only 68 key points are plotted for clear display; for 3D reconstruction, reconstructed shapes are rendered with head light for better view. {Our method offers strong robustness and good performance even in presence of large poses (the 3th, 4th and 5th columns) and occlusions (the 6th, 7th and 8th columns)}. Best viewed in color.}}}
    \label{fig:1}
\end{center}%
}]

\begin{abstract}
\vspace{-0.2cm}
3D face reconstruction from a single 2D image is a challenging problem with broad applications. Recent methods typically  aim to learn a CNN-based 3D face model that  regresses coefficients of 3D Morphable Model (3DMM) from 2D images to render  3D face reconstruction or dense face alignment. However, the shortage of training data with 3D annotations  considerably limits performance of those methods. To alleviate this issue, we propose a novel  2D-assisted self-supervised learning (2DASL) method that can effectively use ``in-the-wild'' 2D face images with noisy landmark information to substantially improve 3D face model  learning. Specifically,  taking  the sparse 2D facial landmarks as   additional information, 2DSAL introduces four novel  self-supervision schemes that view the 2D landmark and  3D landmark prediction as a self-mapping process, including the 2D and 3D landmark self-prediction consistency, cycle-consistency over the 2D landmark prediction and self-critic over the predicted 3DMM coefficients based on landmark predictions. Using these four self-supervision schemes, the 2DASL method significantly relieves  demands on the the conventional paired 2D-to-3D annotations and gives much higher-quality 3D face models without requiring any additional 3D annotations. Experiments on multiple challenging datasets show that our method outperforms state-of-the-arts for both 3D face reconstruction and dense face alignment by a large margin.
\end{abstract}

\vspace{-0.25cm}
\section{Introduction}
3D face reconstruction is an important task in the field of computer vision and graphics. For instance, the recovery of 3D face geometry from a single image can help address many challenges (\textit{e.g.}, large pose and occlusion) for  2D face alignment through dense face alignment \cite{liu2017dense}.
Traditional 3D face reconstruction  methods \cite{amberg2007optimal, paysan20093d} are mainly based on  optimization algorithms, \eg,  iterative closest point \cite{amberg2007optimal}, to obtain coefficients for the 3D Morphable Model (3DMM) model and render the corresponding 3D faces from a single face image \cite{zhu2016face}. However, such methods are usually time-consuming due to the high optimization complexity and suffer from local optimal solution  and bad initialization. Recent works thus propose  to use CNNs to learn to regress the 3DMM coefficients and significantly improve the reconstruction quality  and efficiency.

CNN-based methods \cite{jourabloo2016large, zhu2016face, liu2016joint, liu2017dense, tuan2017regressing, tran2018nonlinear} have achieved remarkable success   in 3D face reconstruction and dense face alignment. However,  obtaining   an accurate 3D face CNN regression model (from input 2D images to 3DMM coefficients) requires a large amount of training faces  with 3D annotations, which are expensive to collect and even not achievable in some cases. Even some 3D face datasets, like 300W-LP \cite{zhu2016face}, are publicly  available, they generally lack diversity in face appearance, expression, occlusions and environment conditions, limiting the generalization performance of resulted 3D face regression models.
A model trained on such datasets  cannot deal well with various  potential cases in-the-wild that are not present in the training examples. Although some recent works bypass the 3DMM parameter regression and use image-to-volume \cite{jackson2017large} or image-to-image \cite{feng2018joint} strategy instead, the ground truths are all still needed and  generated from 3DMM using 300W-LP, still lacking diversity.

In order to overcome the intrinsic limitation of existing 3D face recovery models, we propose a novel learning method that leverages 2D ``in-the-wild'' face images to effectively supervise and facilitate the 3D face model learning. With the method, the trained 3D face model can perform 3D face reconstruction and dense face alignment well.  This is inspired by the observation that a large number of 2D face datasets \cite{bansal2017umdfaces, liu2015deep, koestinger2011annotated, ramanan2012face, sagonas2013300} are available with obtainable 2D landmark  annotations,  that could provide valuable information for 3D model learning, without requiring  new  data with 3D annotations.


Since these 2D images do not have any 3D annotations, it is not straightforward to exploit them in 3D face model learning. We design a novel self-supervised learning method that is able to train a 3D face model with  weak  supervision from 2D images. In particular, the proposed method  takes  the sparse annotated 2D  landmarks  as input and fully leverage the consistency within the 2D-to-2D and 3D-to-3D self-mapping procedure  as supervision. The model should be able to recover 2D landmarks from predicted 3D ones via direct 3D-to-2D projection. Meanwhile, the 3D landmarks predicted from the annotated  and recovered 2D landmarks via the model should be the same. Additionally, our proposed method also exploits cycle-consistency over the 2D landmark predictions, \ie, taking the recovered 2D landmarks as input, the  model should be able to generate 2D landmarks (by projecting its predicted 3D landmarks)  that have small difference with the annotated ones.
By leveraging  these self-supervision derived from 2D face images without 3D annotations, our method could 
substantially improve the quality of learned 3D face regression model, even though there is lack of 3D   samples and no 3D annotations for the 2D samples.
To facilitate the overall learning procedure, our method also exploits self-critic learning. It takes as input both the latent representation and 3DMM coefficients of an face image and learns a critic model to evaluate  the intrinsic consistency between the predicted 3DMM coefficients and the corresponding face image, offering another supervision for 3D face model learning.

Our proposed method is principled, effective and fully exploits available data resources. As shown in Fig.~\ref{fig:1},  our method can produce 3D reconstruction and dense face alignment results with strong  robustness to large poses and occlusions. Our code, models and online demos will be available upon acceptance. Our contributions are summarized as follows:
\begin{itemize}
\setlength\itemsep{0em}
\item We propose a new scheme  that aims to  fully utilize the abundant ``in-the-wild'' 2D face images  to assist 3D face model learning. This is new  and different from most common practices that   pursues   to improve  3D face model by collecting   more data with 3D  annotations for model training.
\item We introduce  a new  method that is able to train 3D face models with 2D face images by self-supervised learning. The devised   multiple forms of self-supervision are effective and data efficient. 
\item We develop a new self-critic   learning based approach which could effectively improve the 3D face model learning procedure and give a better  model, even though the 2D landmark annotations are noisy.
\item Comparison on the AFLW2000-3D and AFLW-LFPA datasets shows that our method achieves excellent performance on both tasks of 3D face reconstruction and dense face alignment.
\end{itemize}

\begin{figure*}[!t]
    \hspace{0.0cm}
    \includegraphics[width=17.5cm, height=7.0cm]{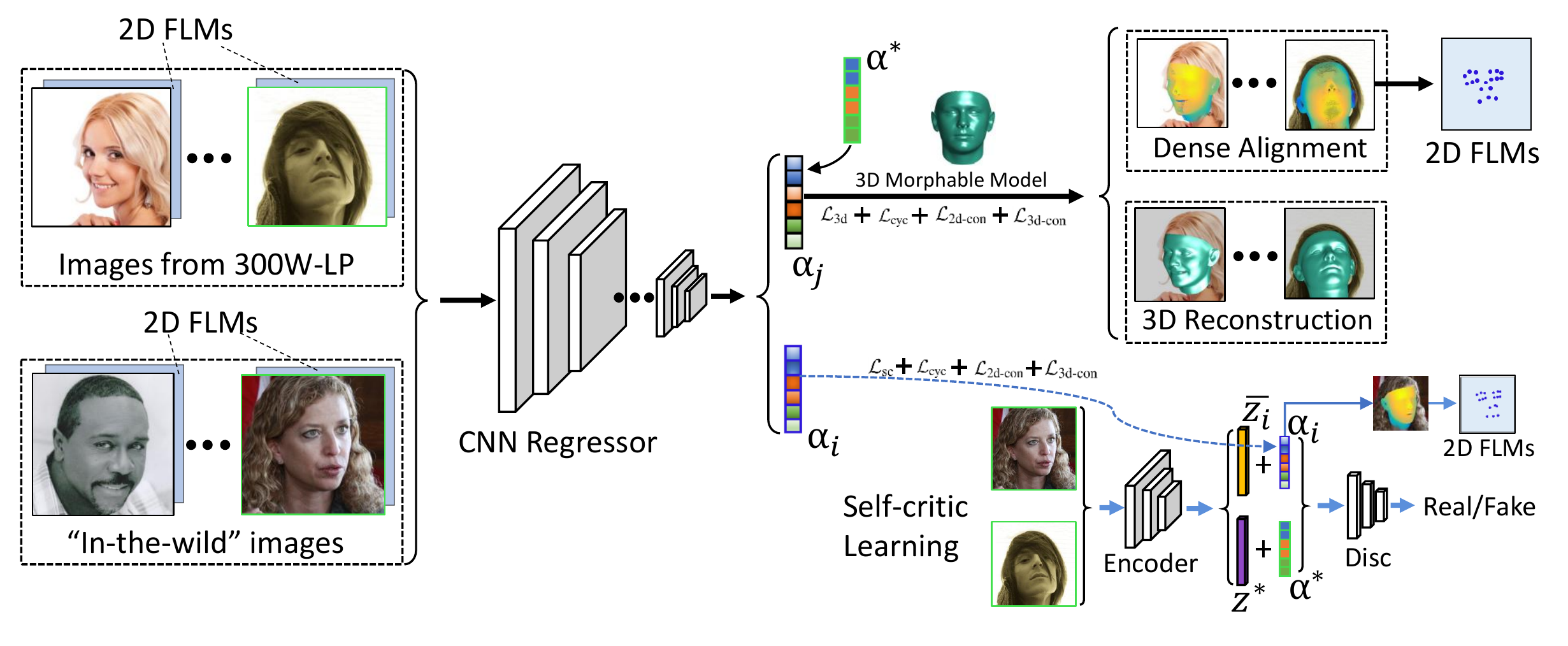}%
   \vspace{-0.55cm}
   \captionsetup{font={small}}
   \caption{The pipeline of our 2DASL. It aims to train a CNN regressor model. The model takes as input the face images with 3D annotations and other images with only 2D Facial Landmark Map (FLM), and predicts coefficients $\bm{\alpha}_{j}$ (for 3D annotated image) and $\bm{\alpha}_{i}$ (for images only with 2D landmarks) for 3DMM for 3D reconstruction and dense alignment. There are dual training  paths. The upper path trains the model through 3D annotation supervision. The bottom path trains the model through self-critic supervision based on the 2D face images. In particular, 2D images are transformed by an encoder to   the latent representation, and the self-critic module evaluates whether the predicted coefficients are consistent with the latent representations, by taking  ground truth pairs as reference. Best viewed in color.}
\label{fig:2}
\end{figure*}

\section{Related work}
\paragraph{3D Face Reconstruction}Various approaches have been proposed to tackle the inherently ill-posed problem of 3D face reconstruction from a single image. In \cite{blanz1999morphable}, Vetter and Blanz observe that both the geometric structure and the texture of human faces can be approximated by a linear combination of orthogonal basis vectors obtained by PCA over 100 male and 100 female identities. Based on this, they propose the 3DMM to represent the shape and texture of a 3D face. After that, large amount of efforts have been proposed to improve 3DMM modeling mechanism. Most of them devote to regressing the 3DMM coefficients by solving the non-linear optimization problem to establish the correspondences of the points between a single face image and the canonical 3D face model, including facial landmarks \cite{zhu2015high,lee2012single,thies2016face2face,cao2014displaced,jeni2015dense,grewe2016fully} and local features \cite{grewe2016fully,huber2015fitting,romdhani2005estimating}. Recently, various attempts have been made to estimate the 3DMM coefficients from a single face image using CNN as a regressor, as opposed to non-linear optimization. In \cite{jourabloo2016large, zhu2016face, richardson20163d,richardson2017learning}, cascaded CNN structures are used to regress the 3DMM coefficients, which are time-consuming due to multi-stage. Besides, end-to-end approaches \cite{dou2017end, tuan2017regressing, jourabloo2015pose} are also proposed to directly estimate the 3DMM coefficients in a holistic manner. More recently, works are proposed to use CNN directly obtain the reconstructed 3D face bypassing the 3DMM coefficients regression. In \cite{jackson2017large}, Jackson \textit{et al.} propose to map the image pixels to a volumetric representation of the 3D facial geometry through CNN-based regression. While their method is not restricted to the 3DMM space any more, it needs a complex network structure and a lot of time to predict the voxel information. In a later work \cite{feng2018joint}, Feng \textit{et al.} store the 3D facial geometry into UV position map and train an image-to-image CNN to directly regress the complete 3D facial structure along with semantic information from a single image.

\vspace{-0.4cm}
\paragraph{Face Alignment}
Traditional 2D face alignment methods aim at locating a sparse set of fiducial facial landmarks. Initial progresses have been made with the classic Active Appearance Model (AAM) \cite{cootes1998active, saragih2007nonlinear, tzimiropoulos2013optimization} and Constrained Local Model (CLM) \cite{asthana2013robust, cristinacce2006feature, saragih2011deformable}. Recently, CNN-based methods \cite{liang2015unconstrained,peng2016recurrent,bulat2017far} have achieved state-of-the-art performance on 2D landmark localization. However, 2D face alignment only regresses visible landmarks on faces, which are unable to address large pose or occlusion situations, where partial face regions are invisible. With the development of this field, 3D face alignment have been proposed, aiming to fit a 3DMM \cite{zhu2016face,mcdonagh2016joint,gou2016shape} or register a 3D facial template \cite{santa20163d,de20163d} to a 2D face image, which makes it possible to deal with the invisible points. The original 3DMM fitting method \cite{blanz2003face} fits the 3D model by minimizing the pixel-wise difference between image and the rendered face model. It is the first method that can address arbitrary poses, which, however, suffers from the one-minute-per-image computational cost. After that, some methods estimate 3DMM coefficients and then project the estimated 3D landmarks onto 2D space, such methods \cite{jourabloo2015pose,cao2014displaced,jeni2015dense,jourabloo2017pose,jourabloo2017pose} could significantly improve the efficiency. Recently, the task of dense face alignment starts to attract more and more research attention, aiming to achieve very dense 3D alignment for large pose face images (including invisible parts). In \cite{liu2017dense}, Liu \textit{et al.} use multi-constraints to train a CNN model, jointly estimating the 3DMM coefficient and provides very dense 3D alignment. \cite{alp2017densereg,yu2017learning} directly learn the correspondence between a 2D face image and a 3D template via a deep CNN, while only visible face-region is considered.

Overall, CNN-based methods have achieved great success in both 3D face reconstruction and dense face alignment. However, they need a huge amount of 3D annotated images for training. Unfortunately, currently face datasets with 3D annotations are very limited. As far as we know, only the 300W-LP \cite{zhu2016face} dataset has been widely used for training. However, the 300W-LP is generated by profiling faces of 300W \cite{sagonas2013300} into larger poses, which is not strictly unconstrained and can not cover all possible scenes in-the-wild.

\section{Proposed method}
In this section we introduce the proposed 2D-Aided Self-supervised Learning (2DASL) method for simultaneous 3D face reconstruction and dense face alignment. We first review the popular 3D morphable model that we adopt to render the 3D faces. Then we explain our method in details, in particular the novel cycle-consistency based  self-supervised learning and the  {self-critic learning}.

\subsection{3D morphable model}
We adopt the 3D morphable model (3DMM) \cite{blanz1999morphable} to recover the 3D facial geometry from a single face image.
The 3DMM renders  3D face {shape} $\bm{S}\in \mathbb{R}^{3N} $ that stores 3D coordinates of $N$ mesh vertices  with {linear combination over} a set of PCA basis. {Following \cite{zhu2016face}}, we use 40  {basis} from the Basel Face Model (BFM) \cite{paysan20093d} to generate the  {face shape} component  and 10 {basis} from the Face Warehouse dataset~\cite{cao2014facewarehouse} to generate the facial expression component. The rendering of a 3D face shape is thus formulated as:
\begin{equation*}
\begin{aligned}
\bm{S} =  \overline{\bm{S}} + \bm{A}_{\text{s}}\bm{\alpha}_{\text{s}} + \bm{A}_{\text{exp}}\bm{\alpha}_{\text{exp}},
\end{aligned}
\end{equation*}
where $\overline{\bm{S}} \in \mathbb{R}^{3 {N}}$ is the mean shape, $\bm{A}_\text{s} \in \mathbb{R}^{3 {N} \times 40}$ is the shape principle basis {trained on the 3D face scans}, $\bm{\alpha}_\text{s} \in \mathbb{R}^{40 }$ is the shape representation coefficient; $\bm{A}_{\text{exp}} \in \mathbb{R}^{3 {N} \times 10}$ is the expression principle basis and $\bm{\alpha}_{\text{exp}} \in \mathbb{R}^{10}$ denotes the corresponding expression coefficient. The target of single-image based 3D face modeling is to predict the coefficients $\bm{\alpha}_{\text{exp}}$ and $\bm{\alpha}_{\text{s}}$ for 3D face rendering from a single 2D image.

After obtaining the  3D face shape $\bm{S}$,  it can be projected onto the 2D image plane with the scale orthographic projection to generate a 2D face from specified viewpoint:
\begin{equation*}
\begin{aligned}
\bm{V} =  f*\bm{Pr}*\bm{\Pi}* \bm{S} + \bm{t},
\end{aligned}
\end{equation*}
where $\bm{V}$ stores the 2D coordinates of the  3D vertices projected  onto the 2D plane, $f$ is the scale factor, $\bm{Pr}$ is the orthographic projection matrix $\left( \begin{array}{ccc} 1 & 0 & 0\\ 0 & 1 & 0\\ \end{array} \right) $, $\bm{\Pi}$ is the projection matrix consisting of 9 parameters, and $\bm{t}$ is the translation vector. Putting them together, we have in total 62 parameters $\alpha = [f,\bm{t},\bm{\Pi},\bm{\alpha}_{\text{s}},\bm{\alpha}_{\text{exp}}]$ to regress for the 3D face regressor model.

\subsection{Model overview}
As illustrated in Fig.~\ref{fig:2}, the proposed 2DASL model contains 3  modules, \ie, a CNN-based  \emph{regressor} that predicts  3DMM coefficients from the input 2D image, an \emph{encoder} that transforms the input image into a  {latent representation}, and a \emph{self-critic} that evaluates   the input (latent representation,  3DMM coefficients) pairs to be  consistent  or  not. 

We use ResNet-50 \cite{he2016deep} to implement the CNN regressor. The encoder contains 6 convolutional layers, each followed by a ReLU and a max pooling layer. The critic consists of 4 fully-connected layers with 512, 1024, 1024 and 1 neurons respectively, followed by a softmax layer to output a score on the consistency degree of the input pair. The CNN regressor takes  a 4-channel tensor as input that  concatenates a 3-channel RGB face image and a 1-channel 2D Facial Landmark Map (FLM). The FLM is a binary-value image, where the locations corresponding to  facial landmarks take the value of  $1$ and  others take the value of $-1$.

Our proposed 2DSAL method trains the model using two sets of images,  \textit{i.e.,} the images with 3DMM ground truth annotations and the 2D face images {with only 2D facial landmark annotations provided  by an off-the-shelf  facial landmark detector \cite{bulat2017far}. The model is trained by minimizing the following one conventional 3D-supervision and four self-supervision   losses.

The first one is the \emph{weighted coefficient prediction loss} $\mathcal{L}_{\text{3d}}$ over the 3D annotated images that measures how accurate the model can predict 3DMM coefficients. The second one is the \emph{2D landmark consistency loss} $\mathcal{L}_{\text{2d-con}}$ that measures how well the predicted 3D face shapes  can recover the  2D landmark locations for the input 2D images. The third one is the \textit{3D landmark consistency loss} $\mathcal{L}_{\text{3d-con}}$. The fourth one is the \emph{cycle consistency loss} $\mathcal{L}_{\text{cyc}}$. {The last one is the \textit{self-critic loss}  $\mathcal{L}_{\text{sc}}$ that estimates the realism of the predicted 3DMM coefficients for 3D face reconstruction, conditioned on the face latent representation.}
Thus the  overall training loss is:
\begin{equation*}
\begin{aligned}
\mathcal{L} = \mathcal{L}_{\text{3d}} + \lambda_1\mathcal{L}_{\text{2d-con}} + \lambda_2\mathcal{L}_{\text{3d-con}} + \lambda_3\mathcal{L}_{\text{cyc}} + \lambda_4\mathcal{L}_{\text{sc}},
\end{aligned}
\end{equation*}
where $\lambda$'s are the weighting coefficients for different losses. The details of these losses are described in the following sections one by one. 

\begin{figure}[t]
    \hspace{0.9cm}
    \includegraphics[width=6cm, height=3cm]{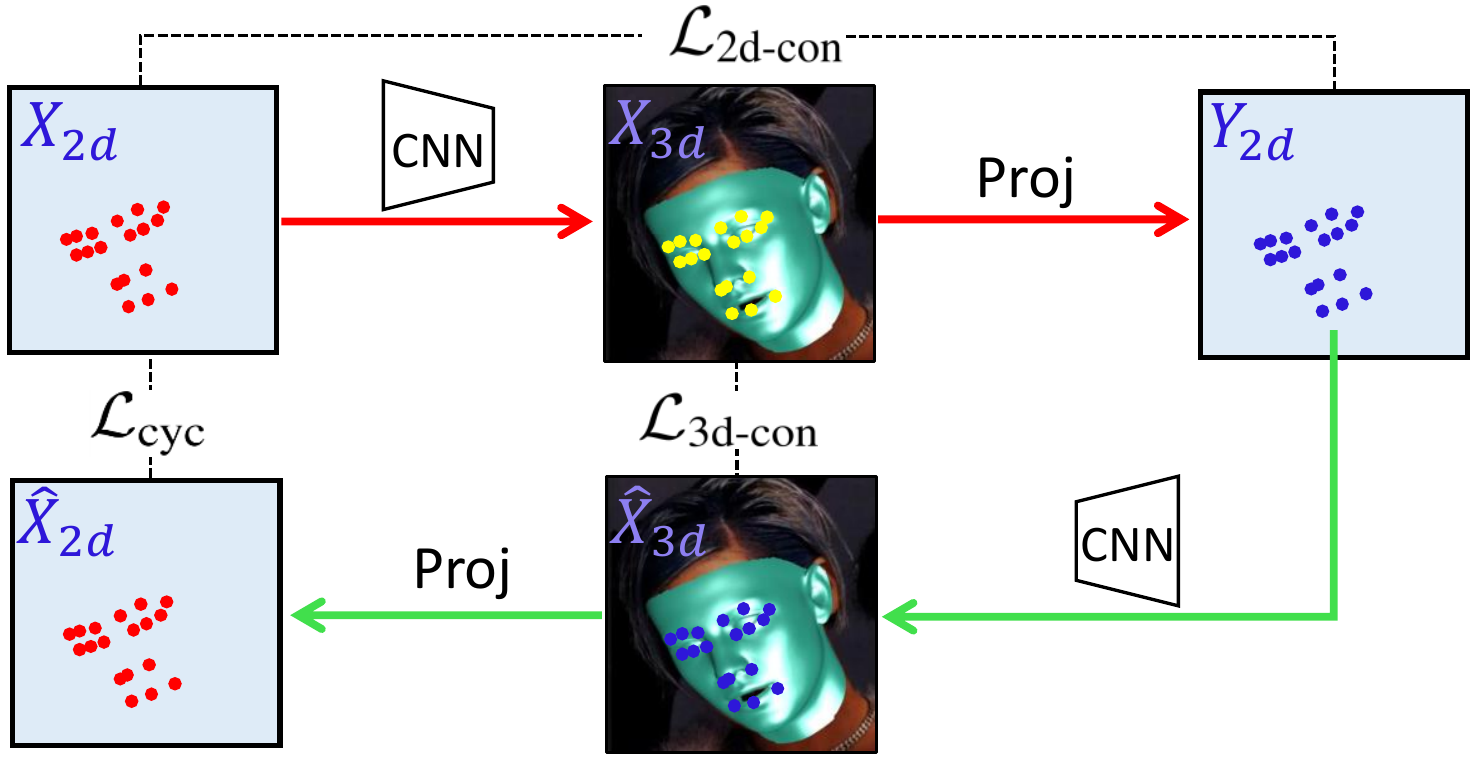}%
   \vspace{-0.2cm}
   \captionsetup{font={small}}
   \caption{{{Illustration on the self-supervision introduced by our 2DSAL for utilizing sparse 2D landmark information. The 2D landmark prediction can be viewed as a self-mapping: $X_{2d} \mapsto Y_{2d}$ (forward training) constrained by $\mathcal{L}_{\text{2d-con}}$. To further supervise the model training, we introduce the $\mathcal{L}_{\text{cyc}}$ by mapping back from $Y_{2d} \mapsto \hat{X}_{2d}$ (backward training). The $\mathcal{L}_{\text{3d-con}}$ is employed to constrain landmarks matching in 3D space during the cycle training. Here $i$ indexes the landmark. Best viewed in color.}}}
\label{fig:3}
\end{figure}
\subsection{Weighted 3DMM coefficient supervision}
Following \cite{zhu2016face}, we deploy the ground truth 3DMM coefficients to supervise the model training where the contribution of each 3DMM coefficient is re-weighted according to their importance. It trains the model to predict closer coefficients $\hat{\alpha}$ to its 3DMM ground truth $\alpha^{*}$.
Instead of calculating the conventional $\ell_2$ loss, we  explicitly consider importance of each coefficient and re-weigh their contribution to the loss computation accordingly. Thus we obtain the {weighted coefficient prediction loss} as follows:
\begin{equation*}
\begin{aligned}
&\mathcal{L}_{\text{3d}} = (\alpha^*-\hat{\alpha})^\top W(\alpha^*-\hat{\alpha}),
\end{aligned}
\end{equation*}
where,
\begin{equation*}
\begin{aligned}
& W = \text{diag}(w_1, \ldots,  w_{62}), \\
& w_i = \frac{1}{\sum_{i} w_i} \|H(\hat{\alpha}_i) - H(\alpha^{*})\|.
\end{aligned}
\end{equation*}
Here $w_i$ indicates  importance of the $i^{th}$ coefficient, computed from how much error it introduces to locations of 2D landmarks after projection. Here $H(\cdot)$ is the sparse landmark projection from rendered 3D shape, $\alpha^{*}$ is the ground truth and $\hat{\alpha}_i$ is the coefficient whose $i^{th}$ element comes from the predicted parameter and the others come from $\alpha^{*}$. With such a reweighting scheme, during training, the CNN model would first focus on learning the coefficients with larger weight (\textit{e.g.}, the ones for rotation and translation). After decreasing their error and consequently their weights, the model will change to optimize the other coefficients (\textit{e.g.}, the ones for shape and expression).

\subsection{2D assisted  self-supervised learning}
To leverage the  2D face images with only annotation of sparse 2D landmark points offered by detector \cite{bulat2017far}, we develop the following  self-supervision scheme that offers three different self-supervision losses, including the 2D landmark consistency loss $\mathcal{L}_{\text{2d-con}}$, the 3D landmark consistency loss $\mathcal{L}_{\text{3d-con}}$ and the cycle-consistency loss $\mathcal{L}_{\text{cyc}}$.

Fig.~\ref{fig:3} gives a systematic overview. The intuition behind this scheme is: if the 3D face estimation model is trained  well, it should present   consistency in the following three aspects. First, the  2D landmarks $Y_{2d}$ recovered  from the predicted   3D landmarks $X_{3d}$ via 3D-2D projection should have small difference with the  input 2D landmarks $X_{2d}$. Second, the predicted 3D landmarks $X_{3d}$ from the input 2D landmarks $X_{2d}$ should be consistent with  the 3D landmarks $\hat{X}_{3d}$ recovered from the predicted 2D landmarks  $Y_{2d}$  by passing it through the same 3D estimation model. Third, the projected $\hat{X}_{2d}$ from $\hat{X}_{3d}$  should be consistent with the original input $X_{2d}$, \ie, forming a consistent cycle.


Thus, we define following two landmark consistency losses in our model correspondingly. The $\mathcal{L}_{\text{3d-con}}$ is formulated as:
\begin{equation*}
\begin{aligned}
\mathcal{L}_{\text{3d-con}} = \sum_{i=1}^{68}{\|x^{3d}_{i}-\hat{x}^{3d}_{i}\|},
\end{aligned}
\end{equation*}
where $x^{3d}_{i}$ is the $i^{th}$ 3D landmark output from the forward pass (see red arrow in Fig.~\ref{fig:3}), $\hat{x}^{3d}_{i}$ is the $i^{th}$ landmark predicted from the backward pass (see green arrow in Fig.~\ref{fig:3}).

For computing the $\mathcal{L}_{\text{2d-con}}$, we first create a weight mask $V = \{v_1, v_2, ..., v_{N}\}$ based the contribution of each point. Since the contour landmarks of a 2D face are inaccurate to represent the corresponding points of 3D face, we discard them and sample 18 landmarks from the 68 2D facial landmarks.
The weight mask is shown in Fig.~\ref{fig:4}.  Here, the mouth center landmark is the midpoint of two mouth corner points. The $\mathcal{L}_{\text{2d-con}}$ is defined as:
\begin{equation*}
\begin{aligned}
\mathcal{L}_{\text{2d-con}} = \sum_{i=1}^{18}{v_i \times \|x^{2d}_{i}-y^{2d}_{i}\| },
\end{aligned}
\end{equation*}
where $x_i^{2d}$ is the $i^{th}$ 2D landmark of the input face, $y_i^{2d}$ is the $i^{th}$ 2D landmark inferred from  the output LMP, and $v_i$ is its corresponding weight. The weight values are specified in Fig.~\ref{fig:4}. We use the following relative  weights  in our experiments: (red points) : (pinky points) : (yellow points) = 4:2:1 that are set empirically.
\begin{figure}[t!]
\center
\vspace{-0.1cm}
\includegraphics[width=3.4cm, height=2.2cm]{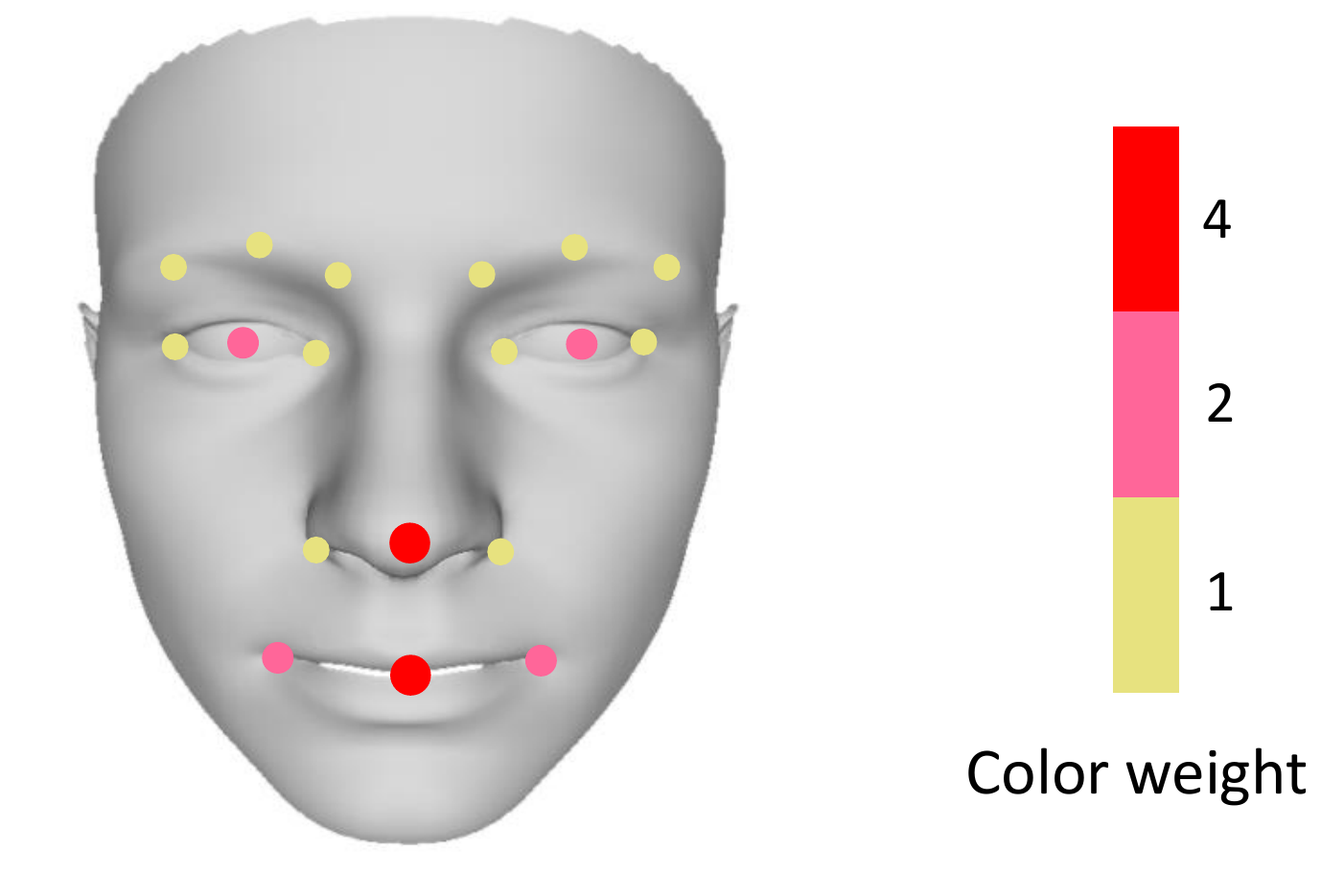}%
\vspace{-0.4cm}
\captionsetup{font={small}}
\caption{{Illustration of the weight mask used for computing $\mathcal{L}_{\text{2d-con}}$. We assign the highest weight to the red points, the medium weight to the pinky points, the yellow points has the lowest weight. Best viewed in color.}}
\label{fig:4}
\end{figure}

\begin{figure*}[t]
\center
\includegraphics[width=17cm, height=4.1cm]{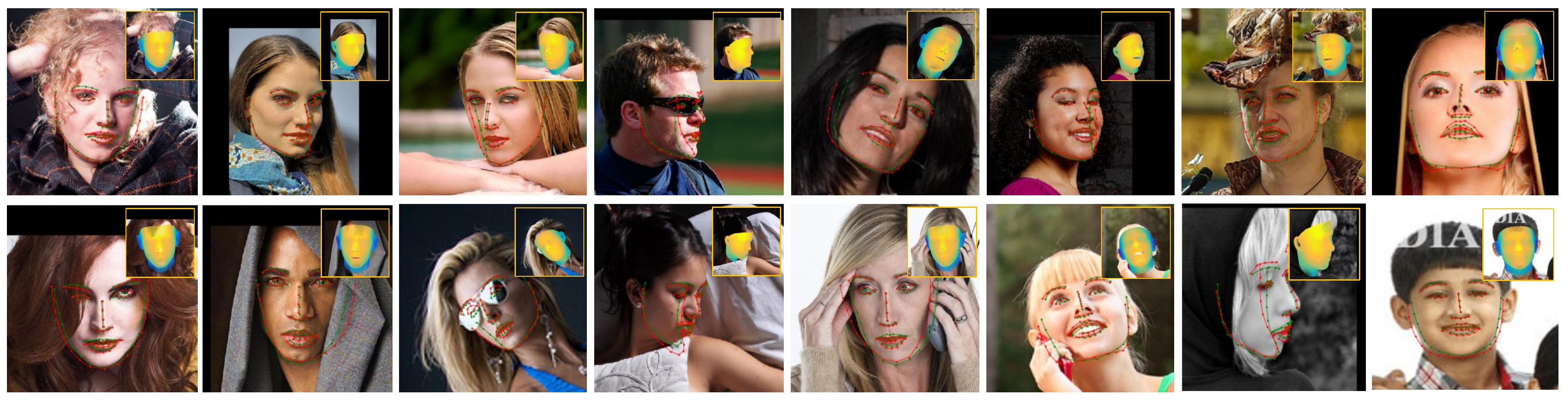}%
\vspace{-0.3cm}
\captionsetup{font={small}}
\caption{Qualitative results on AFLW2000-3D dataset. The predictions by 2DASL show that our predictions are more accurate than ground truth in some cases (only 68 points are plotted to show). Green: landmarks predicted by our 2DASL. Red: ground truth from \cite{zhu2016face}. The thumbnails on the top right corner of each image are the dense alignment results. Best viewed in color.}
\label{fig:5}
\end{figure*}

We model the 2D facial landmarks prediction as a self-mapping process, and denote $F: X_{2d} \rightarrow Y_{2d}$ as the forward mapping, $Q: Y_{2d} \rightarrow X_{2d}$ as the backward mapping. The backward mapping brings the output landmarks $y_i$ back to its original position $x_i$, \textit{i.e.}, $x \rightarrow F(x) \rightarrow Q(F(X)) \approx x$. We constrain this mapping using the \textit{cycle consistency loss}:
\begin{equation*}
\begin{aligned}
\mathcal{L}_{\text{cyc}} = \mathcal{L}_{\text{2d-con}}({x^{2d}}, {\hat{x}^{2d}}),
\end{aligned}
\end{equation*}
where ${x}^{2d}$ are the input 2D facial landmarks, and $\hat{{x}}^{2d}$ are the landmarks output from $Q(F(X))$.

\subsection{Self-critic learning}
We further introduce a self-critic scheme to weakly supervise the model training with the ``in-the-wild'' 2D face images. Given a set of face images $\mathcal{I} = \{I_1, \ldots, I_n\}$ without any 3D annotations and a set of face images $\mathcal{J} = \{(J_1,\alpha_1^*), \ldots, (J_m, \alpha_m^*)\}$ with accurate 3DMM annotations, the CNN regressor model $R: I_i \mapsto \alpha_i$ would output 62 coefficients for each image.  We use another model as the critic  $C(\cdot)$ to evaluate    whether the predicted coefficients are consistent with the input images as the pairs of  $(J_i,\alpha^*_i)$. Since each coefficient is closely related to its corresponding face image, the critic model would learn to distinguish the realism of the coefficients conditioned on the latent representation of the input face images. To this end, we feed the input images to an encoder to obtain the latent representation $z$ and then concatenate with their corresponding 3DMM coefficients as the inputs to the critic $C(\cdot)$. The critic is trained in the same way as the {adversarial learning} by optimizing the following loss:
\begin{equation*}
\begin{aligned}
\mathcal{L}_{\text{sc}} =  \mathbb{E}_{I\in\mathcal{I}}[\log(D([z^*, \alpha^*])) +   \log(1-D([\overline{z}, R(I)]))],
\end{aligned}
\end{equation*}
where $z^{*}$ is the latent representation of a 3D annotated image $J$, $\alpha^{*}$ is the 3DMM ground truth, $I$ is the input ``in-the-wild'' face image, and $\overline{z}$ is its latent representation. The above self-critic loss encourages the model to output 3D faces that lie on the manifold of human faces, and predict landmarks that have the same distribution with the true facial landmarks.

\section{Experiments}
We evaluate 2DASL qualitatively and quantitatively under various settings for 3D face reconstruction and dense face alignment.
\subsection{Training details and datasets}
Our proposed 2DASL is implemented with Pytorch \cite{paszke2017pytorch}. We use SGD optimizer for the CNN regressor with a learning rate beginning at $5\times 10^{-5}$ and decays exponentially, the discriminator uses the Adam as optimizer with the fixed learning rate  $1\times 10^{-4}$. The batch size is set as 32. $\lambda_1$, $\lambda_2$, $\lambda_3$ and $\lambda_4$   are set as 0.005, 0.005, 1 and 0.005 respectively.   We use a two-stage strategy to train our model. In the first stage, we train the model using the overall loss $\mathcal{L}$. In the second stage, we fine-tune our model using the Vertex Distance Cost, following~\cite{zhu2016face}.

The dataset 300W-LP \cite{zhu2016face} is used to train our model. This dataset contains more than 60K face images with annotated 3DMM coefficients. The ``in-the-wild'' face images are all from the UMDFaces dataset \cite{bansal2017umdfaces} that  contains 367,888 still face images for 8,277 subjects. The 2D facial landmarks of all the face images are detected by an advanced 2D facial landmarks detector \cite{bulat2017far}. The input images are cropped to the size $120 \times 120$. We use the test datasets below to evaluate our method:
\vspace{-0.4cm}
\paragraph{AFLW2000-3D \cite{zhu2016face}} is constructed by selecting the first 2000 images from AFLW \cite{koestinger2011annotated}. Each face is annotated with its corresponding 3DMM coefficients and the 68 3D facial landmarks. We use this dataset to evaluate our method on both 3D face reconstruction and dense face alignment.
\vspace{-0.4cm}
\paragraph{AFLW-LFPA \cite{jeni2016first}} is another extension of AFLW. It is constructed by picking images from AFLW according to the poses. It contains 1,299 test images with a balanced distribution of yaw angle. Each image is annotated with 34 facial landmarks. We use this dataset to evaluate performance for the dense face alignment task. The 34 landmarks are used as the ground truth to measure the accuracy of our results.
\begin{figure*}[t]
\center
\hspace{0.2cm}
\includegraphics[width=16.8cm, height=3.8cm]{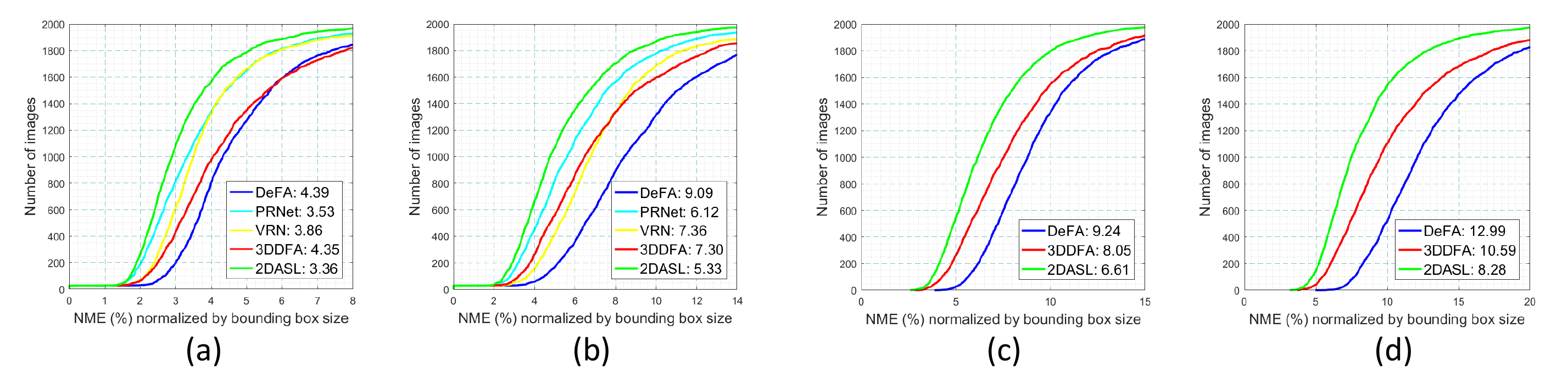}%
\captionsetup{font={small}}
\vspace{-0.3cm}
\caption{Error Distribution Curves (EDC) of face alignment results on AFLW2000-3D. The worst 20 cases of each method are discarded. The horizontal axis are the NME (\%) in ascending order. The vertical axis are the number of images. Evaluation is performed on the 68 2D landmarks (a), 68 3D landmarks (b), all 2D points (c) and all 3D points (d). The mean NME (\%) of each method is shown in the bottom legend.}
\label{fig:6}
\end{figure*}

\vspace{-0.0cm}
\subsection{Dense face alignment}
We first compare the qualitative results from our method and corresponding ground truths in Fig.~\ref{fig:5}. Although all the state-of-the-art methods of dense face alignment conduct evaluation on AFLW2000-3D, the ground truth of AFLW2000-3D is  controversial \cite{bulat2017far, yu2017learning}, since its annotation pipeline is based on the Landmarks Marching method in \cite{zhu2015high}. As can be seen, our results are more accurate than the ground truth in some cases. This is mainly because 2DASL involves a number of the ``in-the-wild'' images for training, enabling the model to perform well in cases even unseen in the 3D annotated training data.

For fair comparison, we adopt the normalized mean error (NME) \cite{zhu2016face} as the metric to evaluate the alignment performance. The NME is the mean square error normalized by face bounding box size. Since some images in AFLW2000-3D contains more than 2 faces, and the face detector sometimes gives the wrong face for evaluation (not the test face with ground truth), leading to high NME. Therefore, we discard the worst 20 cases of each method and only \textit{1,980 images} from AFLW2000-3D are used for evaluation. We evaluate our 2DASL using a sparse set of 68 facial landmarks and also the dense points with both 2D and 3D coordinates, and compare it  with other state-of-the-arts. The 68 sparse facial landmarks can be viewed as sampling from the dense facial points. Since PRNet \cite{feng2018joint} and VRN-Guided \cite{jackson2017large} are not 3DMM based, and the point cloud of these two methods are not corresponding to 3DMM, we only compare with them on the sparse 68 landmarks. The results are shown in Fig.~\ref{fig:6}, where we can see our 2DASL achieves the lowest NME (\%) on the evaluation of both 2D and 3D coordinates among all the methods. For 3DMM-based methods: 3DDFA \cite{zhu2016face} and DeFA \cite{liu2017dense}, our method outperforms them by a large margin on both the 68 spare landmarks and the dense coordinates.

To further investigate   performance of our 2DASL across poses and datasets, we report the NME   of faces with small, medium and large yaw angles on AFLW2000-3D dataset and the mean NME   on both AFLW2000-3D and AFLW-LPFA datasets. The comparison results are shown in Tab. 1. Note that \textit{all the images} from these two datasets are used for evaluation to keep consistent with prior works. The results of the compared method are directly from their published papers. As can be observed, our method achieves the lowest mean NME on both of the two datasets, and the lowest NME   across all poses on AFLW2000-3D. Our 2DASL even performs better than PRNet \cite{feng2018joint}, reducing NME by 0.09 and 0.08 on AFLW2000-3D and AFLW-LFPA, respectively. Especially on large poses (from $60^{\circ}$ to $90^{\circ}$), 2DASL achieves 0.2 lower NME than PRNet. We believe more ``in-the-wild'' face images used for training ensures better performance of 2DASL.

\begin{table}
\small
\begin{center}
\resizebox{0.48 \textwidth}{!}{
\begin{tabular}{|c|c|c|c|c|c|}
\hline
\multirow{2}*{Methods} & \multicolumn{4}{c|}{AFLW2000-3D} & {AFLW-LFPA} \\
\cline{2-6}
        &$0^{\circ}$ to $30^{\circ}$&$30^{\circ}$ to $60^{\circ}$ &$60^{\circ}$ to $90^{\circ}$&Mean&Mean \\
\hline\hline
SDM \cite{mcdonagh2016joint}& 3.67 & 4.94 & 9.67 & 6.12 & - \\
3DDFA \cite{zhu2016face} & 3.78 & 4.54 & 7.93 & 5.42 & - \\
3DDFA + SDM \cite{zhu2016face}& 3.43 & 4.24 & 7.17 & 5.42 & -\\
PAWF \cite{jourabloo2016large}& - & - & - & - & 4.72\\
Yu \textit{et al.} \cite{yu2017learning} & 3.62 & 6.06 & 9.56 & - & - \\
3DSTN \cite{bhagavatula2017faster}& 3.15 & 4.33 & 5.98 & 4.49 & -\\
DeFA \cite{liu2017dense}& - & - & - & 4.50 & 3.86\\
PRNnet \cite{feng2018joint} & 2.75 & 3.51 & 4.61 & 3.62 & 2.93\\
2DASL (ours)& 2.75 & 3.44 & 4.41 & \textbf{3.53} & \textbf{2.85} \\
\hline
\end{tabular}}
\end{center}
\vspace{-0.55cm}
\captionsetup{font={small}}
\caption{Performance comparison on AFLW2000-3D (68 2D landmarks) and AFLW-LFPA (34 2D visible landmarks). The NME (\%) for faces with different yaw angles are reported. The numbers in bold are the best results on each dataset, the lower is the better. ‘‘-'' indicates the corresponding result is unavailable.}
\end{table}

\begin{figure*}[htp]
\center
\hspace{-0.0cm}
\includegraphics[width=17.0cm, height=5.1cm]{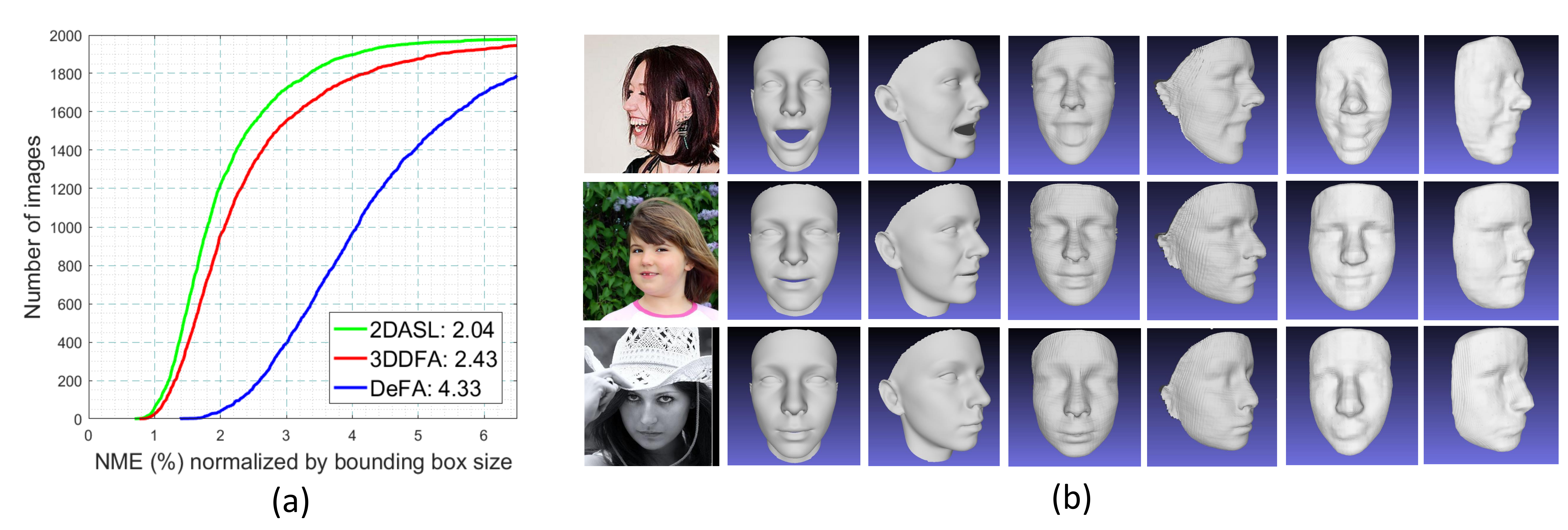}%
\vspace{-0.4cm}
\captionsetup{font={small}}
\caption{(a) EDC of face reconstruction results on AFLW2000-3D dataset. The worst 20 cases of each method are discarded. The mean NME (\%) of each method is shown in the bottom legend. (b) Some 3D reconstruction results of 2DASL (columns 2 \&  3) , PRNet (columns 4 \&  5) and VRN-Guided (columns 6 \& 7). Images of the first column are the original face images.}
\label{fig:7}
\end{figure*}

\vspace{-0.1cm}
\subsection{3D face reconstruction}
In this section, we evaluate our 2DASL on the task of 3D face reconstruction on AFLW2000-3D by comparing with 3DDFA and DeFA. The VRN-Guided and PRNet are not compared because of the mis-match  of point cloud between them and our method. Following   \cite{feng2018joint}, we first employ the Iterative Closest Points (ICP) algorithm to find the corresponding nearest points between the reconstructed 3D face and the ground truth point cloud. We then calculate the NME normalized by the face bounding box size. Fig.~\ref{fig:7} (a) shows the comparison results on AFLW2000-3D. As can be seen, the 3D reconstruction results of 2DASL outperforms 3DDFA by 0.39, and 2.29 for DeFA, which are significant improvements.

We show some visual results of our 2DASL and compare with PRNet and VRN-Guided in Fig.~\ref{fig:7} (b). As can be seen, the reconstructed shape of our 2DASL are more smooth, however, both PRNet and VRN-Guided introduce some artifacts into the reconstructed results, which makes the reconstructed faces look unnaturally.

\vspace{0.2cm}
\subsection{Ablation study}
In this section, we perform ablation study on AFLW2000-3D by evaluating several variants of our model: (1) 2DASL (base), which only takes the RGB images as input without self-supervision and  self-critic supervision; (2) 2DASL (cyc), which takes as input the combination of RGB face images and the corresponding 2D FLMs with self-supervison, however without self-critic supervision; (3) 2DASL (sc), which takes as input the RGB face images only using self-critic  learning. (4) 2DASL (cyc+sc), which contains both self-supervision and self-critic  supervision. For each variant, we use the $\mathcal{L}_{\text{2d-con}}$ with (w/) or without (w/o) weight mask. Therefore, there are in total 6 variants.

The ablation study results are shown in Tab. 2. Adding weights to central points of the facial landmarks reduces the NME   by $0.09$ to $0.23$ on the two stages, respectively. Both self-critic and the self-supervision are effective to improve the performance. If the self-critic learning is not used, the NME increases by 0.04/0.18 for with/without weight mask, respectively. While the self-supervision scheme reduce NME by 0.1 when the weight mask is used, and 0.23 if the weight mask is removed, no significant improvement is observed. The best result is achieved when both these two modules are used. Moreover, in our experiments, we found taking the FLMs as input can accelerate the convergence of training process. Therefore, the first training stage just takes one or two epochs to reach a good model.

\begin{table}[!h]
\small
\begin{center}
\resizebox{0.35\textwidth}{!}{
\begin{tabular}{|l|c|c|c|}
\hline
\multicolumn{2}{|c|}{Variants} & Stage 1 & Stage 2  \\
\hline\hline
\multirow{2}*{2DASL (base)} & w Mask & 4.13 & \textbf{3.77}  \\ \
                            & w/o Mask  & 4.32 & 4.00   \\
\hline \hline
\multirow{2}*{ 2DASL (cyc)} & w Mask & 3.85 & \textbf{3.67} \\
                            & w/o Mask  & 4.03 & 3.79   \\
\hline \hline
\multirow{2}*{ 2DASL (sc)} & w Mask & 3.88 & \textbf{3.73} \\
                            & w/o Mask  & 4,09 & 3.82   \\
\hline \hline
\multirow{2}*{2DASL (cyc+sc)} & w Mask & 3.79 & \textbf{3.53}  \\
                               & w/o Mask  & 3.96 & 3.71  \\
\hline
\end{tabular}}
\end{center}
\vspace{-0.55cm}
\captionsetup{font={small}}
\caption{Ablation study results (NME \%). ``w/ Mask'' means using the weight mask for $\mathcal{L}_{\text{2d-con}}$, while ``w/o Mask'' refers to without weight mask. The numbers in bold are the best results of each variant.}
\vspace{-0.2cm}
\end{table}

\begin{table}[!h]
\small
\begin{center}
\resizebox{0.35\textwidth}{!}{
\begin{tabular}{|c|c|c|c|c|}
\hline
Num. \# ITW & 77,228 & 115,126 & 175,534 & 367,888 \\
\hline\hline
Stage 1 & 4.13 & 4.01 & 3.85 & \textbf{3.79}  \\
Stage 2 & 3.92 & 3.75 & 3.62 & \textbf{3.53}  \\
\hline
\end{tabular}}
\end{center}
\vspace{-0.55cm}
\captionsetup{font={small}}
\caption{The results (NME (\%)) of 2DASL by training with different number of ``in-the-wild'' face images. ``Num. \# ITW'' indicates the number of the ``in-the-wild'' face images used for training. The numbers in bold are the best results of each stage.}
\vspace{-0.1cm}
\end{table}

To explore how the performance is affected by the number of ``in-the-wild'' face images involved in training, we train our model using different numbers. Since the UMDFaces dataset \cite{bansal2017umdfaces} divides the whole dataset into 3 batches, each contains 77,228, 115,126, and 175,534 images respectively. We use the 3 batches and also the whole dataset to train our model. The results are reported in Tab. 3, where we can see the more data that used for aiding training, the lower NME   is achieved by 2DASL.

\vspace{-0.3cm}
\section{Conclusion}
In this paper, we propose a novel 2D-Assisted Self-supervised Learning (2DASL) method for 3D face reconstruction and dense face alignment based on the 3D Morphable face Model. The sparse 2D facial landmarks are taken as input of CNN regressor and learn themselves via 3DMM coefficients regression. To supervise and facilitate the 3D face model learning, we introduce four self-supervision losses, including the self-critic which is employed to weakly supervise the training samples that without 3D annotations. Our 2DASL make the abundant ``in-the-wild'' face images could be used to aid 3D face analysis without any 2D-to-3D supervision. Experiments on two challenging face datasets illustrate the effectiveness of 2DASL on both 3D face reconstruction and dense face alignment by comparing with other state-of-the-art methods.

{\small
\bibliographystyle{ieee}
\bibliography{iccv19_xgtu}

\begin{thebibliography}{10}\itemsep=-1pt

\bibitem{alp2017densereg}
R.~Alp~Guler, G.~Trigeorgis, E.~Antonakos, P.~Snape, S.~Zafeiriou, and
  I.~Kokkinos.
\newblock Densereg: Fully convolutional dense shape regression in-the-wild.
\newblock In {\em CVPR}, pages 6799--6808, 2017.

\bibitem{amberg2007optimal}
B.~Amberg, S.~Romdhani, and T.~Vetter.
\newblock Optimal step nonrigid icp algorithms for surface registration.
\newblock In {\em CVPR}, pages 1--8, 2007.

\bibitem{asthana2013robust}
A.~Asthana, S.~Zafeiriou, S.~Cheng, and M.~Pantic.
\newblock Robust discriminative response map fitting with constrained local
  models.
\newblock In {\em CVPR}, pages 3444--3451, 2013.

\bibitem{bansal2017umdfaces}
A.~Bansal, A.~Nanduri, C.~D. Castillo, R.~Ranjan, and R.~Chellappa.
\newblock Umdfaces: An annotated face dataset for training deep networks.
\newblock In {\em IJCB}, pages 464--473, 2017.

\bibitem{bhagavatula2017faster}
C.~Bhagavatula, C.~Zhu, K.~Luu, and M.~Savvides.
\newblock Faster than real-time facial alignment: A 3d spatial transformer
  network approach in unconstrained poses.
\newblock In {\em ICCV}, pages 3980--3989, 2017.

\bibitem{blanz2003face}
V.~Blanz and T.~Vetter.
\newblock Face recognition based on fitting a 3d morphable model.
\newblock {\em T-PAMI}, 25(9):1063--1074, 2003.

\bibitem{blanz1999morphable}
V.~Blanz, T.~Vetter, et~al.
\newblock A morphable model for the synthesis of 3d faces.
\newblock In {\em SIGGRAPH}, volume~99, pages 187--194, 1999.

\bibitem{bulat2017far}
A.~Bulat and G.~Tzimiropoulos.
\newblock How far are we from solving the 2d \& 3d face alignment problem?(and
  a dataset of 230,000 3d facial landmarks).
\newblock In {\em ICCV}, pages 1021--1030, 2017.

\bibitem{cao2014displaced}
C.~Cao, Q.~Hou, and K.~Zhou.
\newblock Displaced dynamic expression regression for real-time facial tracking
  and animation.
\newblock {\em ACM-TOG}, 33(4):43, 2014.

\bibitem{cao2014facewarehouse}
C.~Cao, Y.~Weng, S.~Zhou, Y.~Tong, and K.~Zhou.
\newblock Facewarehouse: A 3d facial expression database for visual computing.
\newblock {\em T-VCG}, 20(3):413--425, 2014.

\bibitem{cootes1998active}
T.~F. Cootes, G.~J. Edwards, and C.~J. Taylor.
\newblock Active appearance models.
\newblock In {\em ECCV}, pages 484--498, 1998.

\bibitem{cristinacce2006feature}
D.~Cristinacce and T.~F. Cootes.
\newblock Feature detection and tracking with constrained local models.
\newblock In {\em BMVC}, volume~1, page~3, 2006.

\bibitem{de20163d}
F.~H. de~Bittencourt~Zavan, A.~C. Nascimento, L.~P. e~Silva, O.~R. Bellon, and
  L.~Silva.
\newblock 3d face alignment in the wild: A landmark-free, nose-based approach.
\newblock In {\em ECCV}, pages 581--589, 2016.

\bibitem{dou2017end}
P.~Dou, S.~K. Shah, and I.~A. Kakadiaris.
\newblock End-to-end 3d face reconstruction with deep neural networks.
\newblock In {\em CVPR}, pages 5908--5917, 2017.

\bibitem{feng2018joint}
Y.~Feng, F.~Wu, X.~Shao, Y.~Wang, and X.~Zhou.
\newblock Joint 3d face reconstruction and dense alignment with position map
  regression network.
\newblock In {\em ECCV}, pages 534--551, 2018.

\bibitem{gou2016shape}
C.~Gou, Y.~Wu, F.-Y. Wang, and Q.~Ji.
\newblock Shape augmented regression for 3d face alignment.
\newblock In {\em ECCV}, pages 604--615, 2016.

\bibitem{grewe2016fully}
C.~M. Grewe and S.~Zachow.
\newblock Fully automated and highly accurate dense correspondence for facial
  surfaces.
\newblock In {\em ECCV}, pages 552--568, 2016.

\bibitem{he2016deep}
K.~He, X.~Zhang, S.~Ren, and J.~Sun.
\newblock Deep residual learning for image recognition.
\newblock In {\em CVPR}, pages 770--778, 2016.

\bibitem{huber2015fitting}
P.~Huber, Z.-H. Feng, W.~Christmas, J.~Kittler, and M.~R{\"a}tsch.
\newblock Fitting 3d morphable face models using local features.
\newblock In {\em ICIP}, pages 1195--1199, 2015.

\bibitem{jackson2017large}
A.~S. Jackson, A.~Bulat, V.~Argyriou, and G.~Tzimiropoulos.
\newblock Large pose 3d face reconstruction from a single image via direct
  volumetric cnn regression.
\newblock In {\em ICCV}, pages 1031--1039, 2017.

\bibitem{jeni2015dense}
L.~A. Jeni, J.~F. Cohn, and T.~Kanade.
\newblock Dense 3d face alignment from 2d videos in real-time.
\newblock In {\em FG}, volume~1, pages 1--8, 2015.

\bibitem{jeni2016first}
L.~A. Jeni, S.~Tulyakov, L.~Yin, N.~Sebe, and J.~F. Cohn.
\newblock The first 3d face alignment in the wild (3dfaw) challenge.
\newblock In {\em ECCV}, pages 511--520, 2016.

\bibitem{jourabloo2015pose}
A.~Jourabloo and X.~Liu.
\newblock Pose-invariant 3d face alignment.
\newblock In {\em ICCV}, pages 3694--3702, 2015.

\bibitem{jourabloo2016large}
A.~Jourabloo and X.~Liu.
\newblock Large-pose face alignment via cnn-based dense 3d model fitting.
\newblock In {\em CVPR}, pages 4188--4196, 2016.

\bibitem{jourabloo2017pose}
A.~Jourabloo and X.~Liu.
\newblock Pose-invariant face alignment via cnn-based dense 3d model fitting.
\newblock {\em IJCV}, 124(2):187--203, 2017.

\bibitem{koestinger2011annotated}
M.~Koestinger, P.~Wohlhart, P.~M. Roth, and H.~Bischof.
\newblock Annotated facial landmarks in the wild: A large-scale, real-world
  database for facial landmark localization.
\newblock In {\em ICCVW}.

\bibitem{lee2012single}
Y.~J. Lee, S.~J. Lee, K.~R. Park, J.~Jo, and J.~Kim.
\newblock Single view-based 3d face reconstruction robust to self-occlusion.
\newblock {\em EURASIP}, 2012(1):176, 2012.

\bibitem{liang2015unconstrained}
Z.~Liang, S.~Ding, and L.~Lin.
\newblock Unconstrained facial landmark localization with backbone-branches
  fully-convolutional networks.
\newblock {\em arXiv}, 2015.

\bibitem{liu2016joint}
F.~Liu, D.~Zeng, Q.~Zhao, and X.~Liu.
\newblock Joint face alignment and 3d face reconstruction.
\newblock In {\em ECCV}, pages 545--560, 2016.

\bibitem{liu2017dense}
Y.~Liu, A.~Jourabloo, W.~Ren, and X.~Liu.
\newblock Dense face alignment.
\newblock In {\em CVPR}, pages 1619--1628, 2017.

\bibitem{liu2015deep}
Z.~Liu, P.~Luo, X.~Wang, and X.~Tang.
\newblock Deep learning face attributes in the wild.
\newblock In {\em ICCV}, pages 3730--3738, 2015.

\bibitem{mcdonagh2016joint}
J.~McDonagh and G.~Tzimiropoulos.
\newblock Joint face detection and alignment with a deformable hough transform
  model.
\newblock In {\em ECCV}, pages 569--580, 2016.

\bibitem{paszke2017pytorch}
A.~Paszke, S.~Gross, S.~Chintala, and G.~Chanan.
\newblock Pytorch: Tensors and dynamic neural networks in python with strong
  gpu acceleration.
\newblock {\em PyTorch: Tensors and dynamic neural networks in Python with
  strong GPU acceleration}, 2017.

\bibitem{paysan20093d}
P.~Paysan, R.~Knothe, B.~Amberg, S.~Romdhani, and T.~Vetter.
\newblock A 3d face model for pose and illumination invariant face recognition.
\newblock In {\em AVSS}, pages 296--301, 2009.

\bibitem{peng2016recurrent}
X.~Peng, R.~S. Feris, X.~Wang, and D.~N. Metaxas.
\newblock A recurrent encoder-decoder network for sequential face alignment.
\newblock In {\em ECCV}, pages 38--56, 2016.

\bibitem{ramanan2012face}
D.~Ramanan and X.~Zhu.
\newblock Face detection, pose estimation, and landmark localization in the
  wild.
\newblock In {\em CVPR}, pages 2879--2886, 2012.

\bibitem{richardson20163d}
E.~Richardson, M.~Sela, and R.~Kimmel.
\newblock 3d face reconstruction by learning from synthetic data.
\newblock In {\em 3DV}, pages 460--469, 2016.

\bibitem{richardson2017learning}
E.~Richardson, M.~Sela, R.~Or-El, and R.~Kimmel.
\newblock Learning detailed face reconstruction from a single image.
\newblock In {\em CVPR}, pages 1259--1268, 2017.

\bibitem{romdhani2005estimating}
S.~Romdhani and T.~Vetter.
\newblock Estimating 3d shape and texture using pixel intensity, edges,
  specular highlights, texture constraints and a prior.
\newblock In {\em CVPR}.

\bibitem{sagonas2013300}
C.~Sagonas, G.~Tzimiropoulos, S.~Zafeiriou, and M.~Pantic.
\newblock 300 faces in-the-wild challenge: The first facial landmark
  localization challenge.
\newblock In {\em ICCVW}, pages 397--403, 2013.

\bibitem{santa20163d}
Z.~S{\'a}nta and Z.~Kato.
\newblock 3d face alignment without correspondences.
\newblock In {\em ECCV}, pages 521--535, 2016.

\bibitem{saragih2007nonlinear}
J.~Saragih and R.~Goecke.
\newblock A nonlinear discriminative approach to aam fitting.
\newblock In {\em ICCV}, pages 1--8, 2007.

\bibitem{saragih2011deformable}
J.~M. Saragih, S.~Lucey, and J.~F. Cohn.
\newblock Deformable model fitting by regularized landmark mean-shift.
\newblock {\em IJCV}, 91(2):200--215, 2011.

\bibitem{thies2016face2face}
J.~Thies, M.~Zollhofer, M.~Stamminger, C.~Theobalt, and M.~Nie{\ss}ner.
\newblock Face2face: Real-time face capture and reenactment of rgb videos.
\newblock In {\em CVPR}, pages 2387--2395, 2016.

\bibitem{tran2018nonlinear}
L.~Tran and X.~Liu.
\newblock Nonlinear 3d face morphable model.
\newblock In {\em CVPR}, pages 7346--7355, 2018.

\bibitem{tuan2017regressing}
A.~Tuan~Tran, T.~Hassner, I.~Masi, and G.~Medioni.
\newblock Regressing robust and discriminative 3d morphable models with a very
  deep neural network.
\newblock In {\em CVPR}, pages 5163--5172, 2017.

\bibitem{tzimiropoulos2013optimization}
G.~Tzimiropoulos and M.~Pantic.
\newblock Optimization problems for fast aam fitting in-the-wild.
\newblock In {\em ICCV}, pages 593--600, 2013.

\bibitem{yu2017learning}
R.~Yu, S.~Saito, H.~Li, D.~Ceylan, and H.~Li.
\newblock Learning dense facial correspondences in unconstrained images.
\newblock In {\em ICCV}, pages 4723--4732, 2017.

\bibitem{zhu2016face}
X.~Zhu, Z.~Lei, X.~Liu, H.~Shi, and S.~Z. Li.
\newblock Face alignment across large poses: A 3d solution.
\newblock In {\em CVPR}, pages 146--155, 2016.

\bibitem{zhu2015high}
X.~Zhu, Z.~Lei, J.~Yan, D.~Yi, and S.~Z. Li.
\newblock High-fidelity pose and expression normalization for face recognition
  in the wild.
\newblock In {\em CVPR}, pages 787--796, 2015.

\end{thebibliography}
}

\end{document}